%% file: LBP_starAI.tex
\pdfoutput=1
\def\year{2016}\relax
\documentclass[letterpaper]{article}
\usepackage{aaai16}
\usepackage{times}
\usepackage{helvet}
\usepackage{courier}
\usepackage{amsthm}
\usepackage{amsmath}
\usepackage{amsfonts}
\usepackage{graphicx}
\usepackage{algorithm}
\usepackage{enumitem}
\usepackage[noend]{algpseudocode}

\usepackage[font=small,skip=0pt]{caption}

\usepackage[svgnames]{xcolor}
\usepackage{subcaption}
\usepackage{tikz}
\usetikzlibrary{shapes,arrows}
\pgfdeclarelayer{edgelayer}
\pgfdeclarelayer{nodelayer}
\pgfsetlayers{edgelayer,nodelayer,main}
\definecolor{lightgrey}{rgb}{0.922,0.922,0.922}
\definecolor{darkgrey}{rgb}{0.722,0.722,0.722}
\tikzstyle{none}=[inner sep=0pt]
\tikzstyle{clusternode}=[rectangle,fill=lightgrey,draw=Black,align=left,style={inner sep=2,outer sep=2}]
\tikzstyle{variablenode}=[rectangle,fill=lightgrey,draw=Black,style={inner sep=2,outer sep=2}]
\tikzstyle{liftedclusternode}=[rectangle,rounded corners=3pt,fill=darkgrey,draw=Black,line width=0.3mm,align=left,style={inner sep=2,outer sep=2}]
\tikzstyle{liftedvariablenode}=[rectangle,rounded corners=3pt,fill=darkgrey,draw=Black,line width=0.3mm,style={inner sep=2,outer sep=2}]
\tikzstyle{label}=[circle,fill opacity=1.0,align=left,font=\scriptsize]
\tikzstyle{clustertree}=[ellipse,draw=darkgrey]
\tikzstyle{pathedge}=[]
\tikzstyle{regionedge}=[]
\tikzstyle{singletonedge}=[densely dotted]
\tikzstyle{ornode}=[circle,fill=lightgrey,draw=Black]
\tikzstyle{andnode}=[diamond,fill=darkgrey,draw=Black,scale=0.8]
\tikzstyle{gandnode}=[rectangle,fill=darkgrey,draw=Black,scale=0.8]
\tikzset{font={\fontsize{2pt}{1}\selectfont}}

\newcommand{\eat}[1]{}
\newtheorem{theorem}{Theorem}
\newtheorem*{theorem*}{Theorem}
\newenvironment{definition}[1][Definition]{\begin{trivlist}
\item[\hskip \labelsep {\bfseries #1}]}{\end{trivlist}}
\newenvironment{example}[1][Example]{\begin{trivlist}
\item[\hskip \labelsep {\bfseries #1}]}{\end{trivlist}}

\makeatletter
\renewcommand\subsubsection{\@startsection{subsubsection}{3}{\z@}%
                                     {-3.25ex\@plus -1ex \@minus -.2ex}%
                                     {-1.5ex \@plus -.2ex}
                                     {\normalfont\normalsize\bfseries}}
\makeatother
\begin{document}
%
\title{Lifted Region-Based Belief Propagation}

\author{ {\bf David Smith} \\
The University of Texas at Dallas \\
\texttt{dbs014200@utdallas.edu} \\
\And
{\bf Parag Singla} \\
Indian Institute of Technology, Delhi \\
\texttt{parags@cse.iitd.ac.in}
\And
{\bf Vibhav Gogate} \\
The University of Texas at Dallas \\
\texttt{vgogate@hlt.utdallas.edu}
}

\maketitle

\begin{abstract}
Due to the intractable nature of exact lifted inference, research has recently focused on the
discovery of accurate and efficient approximate inference algorithms in Statistical Relational Models
(SRMs), such as Lifted First-Order Belief Propagation. FOBP simulates propositional
factor graph belief propagation without constructing the ground factor graph by identifying
and lifting over redundant message computations. In this work, we propose a generalization of FOBP
called Lifted Generalized Belief Propagation, in which both the 
region structure and the message structure can be lifted. This approach allows more of the
inference to be performed intra-region (in the exact inference step of BP), thereby allowing simulation
of propagation on a graph structure with larger region scopes and  fewer edges, while still maintaining
tractability. We demonstrate that the resulting algorithm converges in fewer iterations to more accurate
results on a variety of SRMs.
\end{abstract}

\section{Introduction}
Statistical relational models (SRMs) have grown in popularity because of their
ability to represent a rich relational structure with underlying uncertainty. However, the discovery of 
general-purpose, fast, and accurate inference algorithms in SRMs has remained elusive.  Exact lifted
inference techniques harness symmetries in the relational structure of SRMs in order to perform
efficient inference, but the involved structure of many real-world domain problems disallow
the use of efficient exact inference. Recent research has focused on discovery of accurate
approximate inference algorithms, such as Lifted Sampling techniques 
\cite{venugopal&gogate12,gogate&al12} and Lifted Belief Propagation
\cite{jaimovich&meshi2012,kersting&al09,broecketal12}.

For example, given a model, Lifted First-Order Belief Propagation (FOBP) \cite{singla&domingos08} simulates
loopy belief propagation on the corresponding propositional factor graph induced by identifying messages that
are provably identical at each iteration of LBP and 'lifting' over them, namely computing them only once and replacing products of identical messages by their appropriate powers (e.g., $\prod_{i=1}^{n} \phi = (\phi)^n$).
The resulting approximation is provably equivalent to the approximation obtained by running
propositional LBP but with a potentially lower time and space complexity.
While FOBP often yields good results in practice, it suffers from the same drawback as LBP;
namely, the accuracy of its approximations depends on the structure of the underlying factor graph.
In general, loopier factor graphs yield poorer approximations.
This problem is exacerbated in relational models, where the underlying factor graphs tend to be densely connected.

Researchers have proposed a myriad of LBP variants in order to improve the algorithm's efficacy. One significant
line of research has focused on the observation that factor graphs with fewer loops tend to converge more often
and to better approximations (for example, on tree-structured factor graphs BP yields
exact answers, and that factor graphs with a single loop \emph{always} converge, although to possibly
erroneous approximations \cite{weiss2000}). One way to reduce the number of loops is to reduce the
number of edges in the message passing structure; therefore a large-class of algorithms
specify some generalization of the factor graph structure that allows factors to be
clustered together into regions (e.g. \cite{yedidiaetal2005,dechteretal2002}). These algorithms allow for the
exchange of  cheap, approximate inference (i.e. inter-cluster message passing) for expensive, exact inference
(i.e. intra-cluster variable elimination). The resulting schemes allow the user to trade algorithmic complexity for
more likely convergence and better approximation accuracy.

We propose a generalized belief propagation scheme for SRMs. The scheme employs exact lifted inference
rules to compactly encode the potential structure at each region, thus admitting regions with much larger factor
and variable sets than possible with propositional schemes. 
Our scheme harnesses the symmetric nature of relational models in order to pass joint messages
over groups of exchangeable variables. In conjunction as well as offloading the approximate,
inter-cluster inference step of LBP (message passing) into the exact, intra-cluster step of LBP (sum-product inference)
whenever efficient, allowing the simulation of propagation on region graphs with larger region
scopes and fewer edges while still maintaining tractability. We demonstrate that the resulting
algorithm converges in fewer iterations to more accurate results on a variety of relational models.

\section{Background}
\subsection{Markov Logic}
Statistical relational modeling languages combine graphical models with elements of
first-order logic, by defining template features that apply to whole classes of objects at once.
One such simple and powerful language is Markov logic \cite{richardson&domingos2006}. We formally
define a Markov Logic Network as follows:

\begin{definition}
A \emph{Markov Logic Network} (MLN) $M$ is a pair $\langle F,C \rangle$, in which $F$ is a 
set of weighted clauses, $\{ \langle f_1, w_1 \rangle,\ldots, \langle f_n, w_n \rangle \}$,
where $f_i$ is a first order clause (all logical variables in $f_i$ are assumed to be universally quantified and standardized apart for simplicity) and $w_i \in \mathbb{R}$ is its corresponding weight, and
$C$ is a list of constraints over the logical variables of each $f_i$.
We adopt the constraint language similar to that presented in
\cite{mittaletal2015}, in which each constraint is either a domain constraint (i.e. $x \in \tau_i$,
where $\tau_i$ is an ordered set of constants or objects $\{c_1,\ldots, c_n\}$ called the domain of $x$),
an equality constraint (i.e. $x = y$), or an inequality constraint (i.e. $x \neq y$).
\end{definition}

Let $V=lvars(F)$, the set of all logical variables in $F$. Then the tuple $\langle V, C \rangle$ defines a constraint satisfaction problem.
Let $\Theta$ be the set of solutions to $\langle V, C \rangle$.
Then $\{R\theta \mid R \in F, \theta \in \Theta \}$ is the set of \emph{ground atoms} of $M$, and
$\{ f_i\theta \mid f_i \in F, \theta \in \Theta \}$ is the set of \emph{ground formulas} of $M$. For example, the first-order clause $\forall x \forall y \;S(x) \vee \neg T(y)$ given the constraint $x\neq y$ and constants $\{a_1,a_2\}$ yields the following two ground features: $S(a_1) \vee \neg T(a_2)$ and $S(a_2) \vee \neg T(a_1)$.
Every MLN defines a Markov network with one node per ground atom and one feature per ground
formula. The weight of a feature is the weight of the first-order clause that originated it. The probability
of a state $x$ in such a network is given by $P(x) = \frac{1}{Z} \exp( \sum_i w_i g_i(x))$, where $w_i$ is the
weight of the $i$-th (ground) feature, $g_i(x) = 1$ if the $i$-th feature is true in $x$, and $0$ otherwise. 

\subsection{Generalized Belief Propagation}
\label{GBP}
Loopy Belief propagation \cite{pearl88} is an approximate inference procedure
for graphical models. Given a model, the algorithm operates by iteratively passing messages
 between adjacent nodes on the corresponding factor graph until marginal beliefs converge for all
 variables in the model (or a bound on the number of iterations is reached). Generalized Belief Propagation \cite{yedidiaetal2005} is a generalization of
 the LBP algorithm that operates on an underlying graph structure called a region graph.
\begin{definition}
Given a PGM $P = \langle X,F \rangle$, where $X$ is a set of random variables and $F$ is a set of
factors, a \emph{region graph} is a labeled, directed graph $G = (V, E, L)$,
in which each vertex $v \in V$ (corresponding to a region) is labeled with a subset of $X$ and a
subset of $F$. We denote the label of vertex $v$ by $l(v) \in L$. A directed edge $e \in E$ may exist
pointing from vertex $v_p$ to vertex $v_c$ if $l(v_c)$ is a subset of $l(v_p)$. 
\end{definition}
In the canonical message passing formulation (called the parent-to-child algorithm),
each region $R$ has a belief $b_R(x_R)$ given by:
\begin{multline}
b_R(x_R) =\prod_{a \in a_R} f_a(x_a)\left( \prod_{P \in P(R)} m_{P \rightarrow R}(x_R) \right)
\\
\left( \prod_{D \in D(R)} \prod_{ P' \in P(D) \setminus \mathcal{E}(R)} m_{P' \rightarrow D} (x_D) \right)
\end{multline}

\normalsize
Here $P(R)$ is the set of regions that are parents to region $R$, $D(R)$ is the set of all
regions that are descendants of region $R$, $\mathcal{E}(R) = R \cup D(R)$ is the set of all regions
that are descendants of $R$ and also region $R$ itself, and $P(D) \setminus \mathcal{E}(R)$ is the set of all
regions that are parents of region $D$ except for region $R$ itself or those regions
that are also descendants of region $R$. 
The message-update rule is derived by insisting on equality between the joint distributions between
adjacent nodes.

\section{Exchangeable Normal Form}
\label{enf}
Our proposed Lifted Generalized Belief Propagation (LGBP) algorithm relies on the exchangeable nature of the ground formulas associated with a lifted
formula in order to send and receive compact messages over large groups of variables. As such,
the algorithm requires that the input MLN be preprocessed into a format that facilitates construction of
these messages. We call it exchangeable normal form, defined formally below:
\begin{definition}
Let MLN $M=\langle F, C \rangle.$ Let $G_i$ be the set of \emph{ground formulas} associated with
formula $f_i \in F$. $M$ is said to be in \emph{exchangeable normal form} if and only if $\forall g_j,g_k
\in G_i$,  the joint distribution $P(Vars(g_j))$ equals $P(Vars(g_k))$ subject to renaming of the random variables, where $Vars(g_i)$ is the set of propositional (random) variables in $g_i$.
\end{definition}
\begin{example}
Consider the MLN $M$ consisting of the single formula:
\scriptsize
\begin{gather*}
\langle S(x) \vee \neg S(y) \vee \neg F(x,y), w \rangle \{ x,y \in \{a_1,a_2\} \}
\end{gather*}
\normalsize
$M$ is not in exchangeable normal form. The ground formulas in which $x=y$ can have a different
distribution than those in which $x \neq y$. To see why, note that if $x=y$ the ground formula
becomes a tautology, whereas if $x \neq y$, it does not. However, we can rewrite the formula of
$M$ as $M'$, in which the formula is \textit{shattered} into two formulas with associated constraints.
\scriptsize
\begin{gather*}
\langle S(x_1) \vee \neg S(y_1) \vee \neg F_{x=y}(x_1,y_1), w \rangle \\
\langle S(x_2) \vee \neg S(y_2) \vee \neg F_{x\neq y}(x_2,y_2), w \rangle \\
\{ x_1,x_2,y_1,y_2 \in \{a_1,a_2\}, x_1 = y_1,x_2 \neq y_2\}
\end{gather*}
 \normalsize
$M'$ is in \emph{exchangeable normal form}.
\end{example}

\section{Lifted Inference}

Lifted inference is a collection of techniques that exploit the symmetries in  
graphical models in order to efficiently compute the partition function
(via sum-product based inference). Since its introduction \cite{poole03},
researchers have developed a variety of algorithms for performing exact
lifted inference (e.g. \cite{braz07,gogate&domingos11b,broeck&al11,smith&gogate2015}).
Each of these algorithms rely on a handful of \emph{lifting rules} that dictate when and how
to perform inference efficiently. We discuss two rules that are common to popular algorithms.

\begin{definition}
({\bf Lifted Sum.}) Given a model $M$ with set of exchangeable random variables $X$,
where $\lvert X \rvert = n, Z(M)=
\sum_{k=0}^n {n \choose k}Z(M \lvert \{x_1,\ldots,x_k\}=T,\{x_{k+1},\ldots,x_{n}\}=F)$
\end{definition}
\begin{definition}
({\bf Lifted Product.})
Given a model $M$ that is decomposable into a collection of independent subproblems
$\{{\bf M_1}, \ldots, {\bf M_n}\},$ where each subproblem $M_i \in {\bf M_i}$ is identical,
the partition function of $M, Z(M)=\prod_{i=1}^n Z(M_i)^{\lvert {\bf M_i} \rvert}$
\end{definition}

Exact lifted inference can be applied to any PGM, but it is
particularly effective on templated models (such as MLNs) because (1) sets of independent and
identical subproblems \emph{and} (2) sets of exchangeable random variables can often be readily
identified from the template structure.
We can view the heuristic decisions as to which lifting rules to apply during execution on model $M$
as a partially ordered set. Further, because unordered pairs of elements represent the roots of
independent subproblems, the ordering defines a rooted tree, which we call a \emph{lifted factorization}.
\begin{definition}
A \emph{lifted factorization} for model $M$ is a rooted, labeled tree $E_M=\langle V, E \rangle$, in
which:
\begin{enumerate}[topsep=0pt,itemsep=-1ex,partopsep=1ex,parsep=1ex,leftmargin=0.5cm]
\item each vertex $v \in V$ is labeled by a $k$-arity predicate $R(i_1,\ldots,i_k),$ where $\{i_1,\ldots,i_k\} \in 
\{ C,D,G \}$, where:
\begin{enumerate}[topsep=-10pt,itemsep=-1ex,partopsep=1ex,parsep=1ex,leftmargin=0.5cm]
\item $i_j=C$ indicates that the inference
algorithm performs the lifted sum operation over the set of exchangeable random variables
represented by $R\theta,$ where $\theta = \{ x_1 = c_1,x_k = c_k, x_j \in D_{x_j}\}$. 
\item $i_j=D$ indicates that the inference algorithm has decomposed over the
set of logical variables appearing at position $j$ in predicate $R$ in $M$ (lifted product rule), and
\item $i_j=G$ indicates that the inference algorithm grounds the set of logical variables
appearing at position $j$ in predicate $R$ in $M$.
\end{enumerate}
\item each edge $e \in E$ is labeled by a (possibly empty) set of logical variables $X$
that decompose the subproblem represented by the tree below into identical subproblems.
\end{enumerate}
\end{definition}

\begin{figure}[t]
\center
\input{figExTree1.tikz}
\caption{Two lifted factorizations for the 
MLN $R(x) \vee S(y)$ with $\{ x \in \{1,2 \}, y \in \{1,2\} \}$, and their lifted inference representations.}
\label{executiontree}
\end{figure}
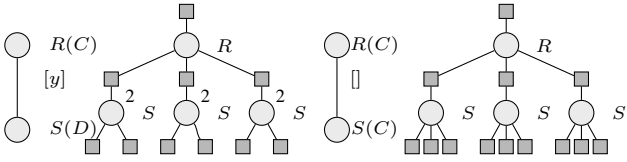

A lifted factorization is \emph{valid} for model $M$ if the application of each
inference rule over the subtree rooted at each node is valid (i.e. meets the preconditions of the rule).
All valid lifted factorizations for $M$ are correct in that they return the same partition function. However, 
each choice encodes a different factorization of the (unnormalized) joint probability
distribution. Therefore, some lifted factorizations yield more efficient inference than others. 
Further, the joint marginal probability distribution of a set of random variables is only (efficiently) available
if they occur on the same path from root to leaf. Hence, different factorizations admit efficient access
to the joint distribution over different sets of random variables.

\begin{example}
Consider the MLN $M = R(x) \vee S(y)$ with $\{ x \in \{1,2 \}, y \in \{1,2\} \}$.
Figure \ref{executiontree} (left) shows a possible lifted factorization for $M$, which applies the Lifted
Sum Rule to $R$, then applies the Lifted Product Rule to $\{y\}$, then applies the Lifted Sum Rule to
a single grounding of $S$. This lifted factorization yields a search space with $6$ leaves, which
admits efficient access to the joint marginal distribution over sets $\{ R(1), R(2), S(1) \}$ or
$\{ R(1), R(2), S(2) \}$ (which are equivalent up to a renaming of $S$), but not over the full joint
distribution $\{ R(1), R(2), S(1), S(2) \}$. Figure \ref{executiontree} (right) does not apply the lifted
product rule, yielding a (larger) lifted search space with $9$ leaves, which admits efficient access
to the joint marginal distribution over all subsets of the random variables $\{ R(1), R(2), S(1), S(2) \}$.
\end{example}

\begin{definition}
Given a MLN $M$ with ground atoms $A_{M}$ and an associated
valid lifted factorization $E$, define $JD(M,E) = \{ V \mid V \subseteq A_{M}, P(V)$ can be accessed
efficiently under lifted factorization $E \}$.
\end{definition}

\section{Lifted Generalized Belief Propagation}

\begin{figure}
\centering
\input{fig1.tikz}
\caption{Three types of simulated region graphs for the model $R(x) \vee S(y), R(x) \vee T(z),$
with domain sizes $\Delta_x = \Delta_y = \Delta_z = \{1 \ldots 10\}$.
Light grey rectangles represent ground factors. Light grey circles represent ground atoms.
Dark grey rectangle represent lifted factors. Dark grey circles represent lifted atoms.
}
\label{simulatedregiongraphs}
\end{figure}
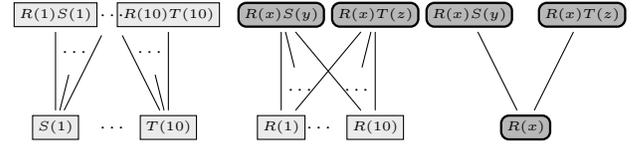

Given a model $M$, FOBP \cite{singla&domingos08} takes advantage of redundant messages in order to
\emph{simulate} the message passing procedure on the factor graph of $M$ without explicitly constructing
the factor graph. We refer to this kind of lifting operation as \emph{message-based} lifting. Our new scheme, Lifted Generalized
Belief Propagation (LGBP) improves on this algorithm in two ways. First, LGBP harnesses lifted inference
rules in order to compactly represent large sets of factors and variables within a cluster whenever it is efficient.
We refer to this kind of lifting operation as \emph{region-based} lifting. Second, wherever it is possible LGBP
uses a lifted representation of the messages themselves; this representation allows message passing over the 
joint distribution of collections of exchangeable atoms rather than over multiple copies of singleton atoms.

\begin{example}
Figure \ref{simulatedregiongraphs} depicts three variants of simulated region graphs for the MLN
$R(x) \vee S(y), R(x) \vee T(z)$, with constraint set $x \in \{a_1,\ldots a_{10}\},y \in \{b_1,\ldots b_{10}\},
z \in \{ c_1, \ldots, c_{10}\}$.
Figure \ref{simulatedregiongraphs} (left) depicts the propositional factor graph (which FOBP simulates).
Figure \ref{simulatedregiongraphs} (middle) depicts the region graph in which all factors are lifted
(via \emph{region-based} lifting), but messages are still passed over ground variables
(via \emph{message-based} lifting). Figure \ref{simulatedregiongraphs} (right) depicts a region graph
in which the factors and messages are lifted (i.e. all groundings of each formula in the MLN appear
within the same cluster, and the clusters communicate through a single message containing the
joint distribution over $\{ R(1) \ldots R(10) \}$). In this case the simulated region graph is a tree;
hence, inference is exact.
\end{example}

In particular, if the complexity of propositional region graph BP is $O(n\exp(w))$ where $n$ is the number of messages and $w$ is the maximum number of random variables in each ground region (the complexity of inference in each region is exponential in $w$), message-based lifting reduces $n$ while region-based lifting reduces $w$. 

\subsection{Lifted Region Graphs}
Propositional GBP operates on a region graph. A region graph is a directed, acyclic,
labeled graph, in which each label defines (1) the scope of variables at a region and
(2) the set of potential functions at a region. FOBP operates on a lifted network, which is
a template that defines a ground factor graph upon which LBP is simulated.
LGBP requires a structure which combines these two definitions; it operates on a templated
graph structure that encodes additional information about the lifting operations occurring
both within a region and between adjacent regions.

\subsubsection{Lifted Region Nodes}
\label{liftedregionnodes}
A lifted region node is a template that defines the lifted inference procedure over a set of random variables.
We begin with some definitions:
\begin{definition}
Let MLN $M = \langle F, C\rangle$. Let $V = lvars(F)$. Let $V_g \subseteq V$. 
Let $\Theta$ be the set of consistent evaluations of the CSP $\langle V,C \rangle$.
Define $\Theta_{V_g}$ as the restriction of $\Theta$ to the variables in $V_g$, i.e.
$\{ \theta_{V_g} \mid \theta \in \Theta \}$. A \emph{partial grounding}
of $M$ with respect to $V_g$ is the MLN $M' = \langle F, C \cup \theta_{v_g} \rangle.$
\end{definition}

\begin{theorem}
\label{exchangeableMLNs}
Let MLN $M=\langle F, C \rangle$ be in exchangeable normal form. Let $V_g \subseteq lvars(F)$.
Then every partial grounding of $M$ with respect to $V_g$ represents an identical joint probability
distribution up to a renaming of variables.
\end{theorem}

Theorem \ref{exchangeableMLNs} follows immediately from the definition of Exchangeable Normal Form.
In propositional GBP, each region $R$ is labeled by (1) a set of factors $F$,
and (2) a set of random variables $X$ such that $\forall \phi \in F, Scope(\phi) \subseteq X.$ At each
lifted region $r$, LGBP requires additional information about (1) how the joint distribution at $r$ is
encoded (to exploit region based symmetries), and (2) how the node is templated in the ground
region graph (to exploit message based symmetries).

\begin{definition}
A \emph{Lifted Region} is a triple $r = \langle M_r,$ $V_g,$ $E_{r_g} \rangle$, where $M_r=\langle F_r, C_r \rangle$
is a MLN in exchangeable normal form,
$V_g \subseteq lvars(F_r)$,
$M_{r_g}$ is a partial grounding of $M_r$ with respect to $V_g$,
and $E_{r_g}$ is a lifted factorization such that $\forall$ ground formulas $g$ of $M_{r_g}$,
$\exists V \in JD(M_{r_g},E_{r_g})$ such that $Atoms(g) \subseteq V$.
\end{definition}

For notational convenience, we assume that the set of formula at each lifted region contains all
the predicates appearing in $E_{r_g}$. These predicates can always be added as singleton
formula with zero weights.
If $M_{r_g}$ is the set of partial groundings of $M_r$ with respect to $V_g$, then the lifted region $r$
represents $\lvert M_{r_g} \rvert$ ground regions in the propositional region graph that LGBP
simulates at inference time.
Thus, the sets $V_g$ and $V \setminus V_g$ represent the sets of logical variables over which
we perform inference via \emph{message-based} lifting and \emph{region-based} lifting respectively.

\subsubsection{Lifted Region Edges}
In LGBP, the distribution at each region is represented by some factorization $E_{r_g}$ rather than 
as a flat table (as in proposition GBP).
This additional structure complicates the parent-child relationship in two ways. First, it is only possible to extract
messages over collections of ground atoms $JD(E_{r_g}).$ Second, whenever possible, the
joint marginal over the group of exchangeable variables of the form $R(x_1,\ldots,x_k)$
is 'lifted' into the space of $O(n)$ parameters. These `lifted` messages are only compatible if
the encoding is the same in each region. Formally:
\begin{definition}
A lifted region $r_p = \langle M_{r_p},$ $V_{p_g},$ $E_{r_{p_g}} \rangle$ is \emph{marginal compatible}
with lifted region $r_c = \langle M_{r_c},$ $V_{c_g},$ $E_{r_{c_g}} \rangle$
 on lifted atom $R$
if and only if (1) $R(p_1,\ldots, p_k) \in E_{r_{p_g}}$, (2) $R(c_1,\ldots, c_k) \in E_{r_{c_g}}$, and
(3) $\forall i \in \{1\ldots k\}, c_i = C \rightarrow p_i = C$.
\end{definition}
\begin{definition}
A lifted region $r_p = \langle M_{r_p},$ $V_{p_g},$ $E_{r_{p_g}} \rangle$ is \emph{message compatible}
with lifted region $r_c = \langle M_{r_c},$ $V_{c_g},$ $E_{r_{c_g}} \rangle$
if and only if (1) $\forall R \in E_{r_{c_g}}, r_p$ and $r_c$ are \emph{marginal compatible} on $R$,
(2) $E_{r_{c_g}}$ is a path graph, and (3) the set of lifted atoms $\{ R \mid R \in E_{r_{c_g}} \}$ all
occur on a single path in $E_{r_{p_g}}$.
\end{definition}
\begin{definition}
A \emph{lifted region edge} is a pair $\langle r_p, r_c \rangle$, where $r_p$ is a parent region,
$r_c$ is a child region, and $r_p$ is \emph{message compatible} with $r_c$.
\end{definition}
The above definitions insure that for $r_p$ and $r_c$ to pass messages, all of the random
variables represented by a grounding of $r_c$ are jointly accessible in the factorization of $r_c$.

\subsubsection{Lifted Region Graph Definition}
\begin{definition}
A \emph{Lifted Region Graph} is a pair $\langle R, E \rangle$, where $R$ is a set of lifted regions and
$E$ is a set of lifted edges.
\end{definition}

\begin{figure}[t!]
\center
\input{figRST.tikz}
\input{figRSTground.tikz}
\caption{A lifted region graph for the MLN $\{ R(x) \vee S(y), S(y) \vee T(z), R(x) \vee T(z)\}$, and it corresponding simulated region graph }
\label{rstRG}
\end{figure}
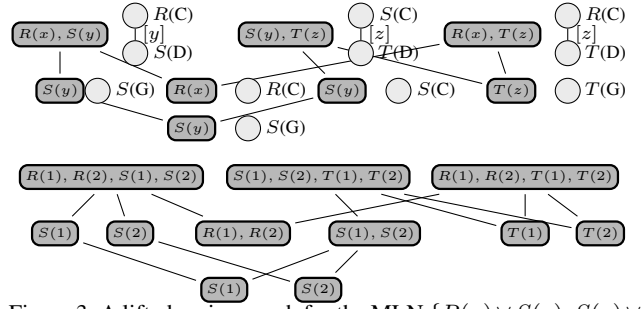

\begin{example}
Figure \ref{rstRG} represents a possible Lifted Region Graph for the MLN $\{ R(x) \vee S(y),
S(y) \vee T(z), R(x) \vee T(z)\}$. Each region represents all the groundings of a single formula from the MLN;
each formula is factorized by counting over the first predicate and decomposing over the second predicate.
Each occurrence of lifted atom $R$ is counted over; therefore, regions containing $R$ communicate via
a joint message over all groundings of $R$. Each occurrence of lifted atom $T$ is decomposed upon; hence
the factorization at each region does not have access to the joint marginal over $T$. Messages are passed
over each grounding of $T$. Lifted atom $S$ is counted over in one region and decomposed over in
another region. These message formats are incompatible.
We reconcile the incompatibility by defaulting to communication via a third level region node connecting
the incompatible $S$ nodes via ground messages.
\end{example}

\subsection{The Simulated Region Graph}
\label{simulatedregiongraph}

Each lifted region graph $R_l$ corresponds to a unique ground region graph $R_g$ upon which the
LGBP algorithm simulates propagation.
Given a lifted region graph $R_l$, we can construct the corresponding ground region graph $R_g$
in a straightforward manner.

For each lifted region $r_i = \langle \langle F_i,C_i \rangle, V_{ig}, E_{r_ig} \rangle \in R_L$, construct the set of vertices
and labels for each ground region it represents. $r_i$ represents a ground region for each assignment
to all variables in $V_{ig}$ consistent with constraint set $C_i$.
Let $\Theta_{r_i} = Sols(\langle V_{ig}, C_i \rangle)$. Let $\theta_{r_{ij}} \in \Theta_{r_i}$ be the partial
groundings of $r_i$ with respect to variable set $V_{ig}$.
Let $V_{i} = lvars(F_i)$.
Define $Labels(r_i) = \{ 
\langle A_g(\theta_{r_{ij}}), F_g((\theta_{r_{ij}})\rangle \mid \theta_{r_{ij}} \in \Theta_{r_{i}} \}$,
where
$A_g(\theta_{r_{ij}})= \{ R\theta \lvert R \in F_i,\theta \in Sols(\langle V_{i}, C_{i} \cup \theta_{r_{ij}} \rangle)$
is the set of ground atoms of $M_{r_i}$ corresponding to $\theta_{r_{ij}}$,
and $F_g((\theta_{r_{ij}})= \{ f_i\theta \mid f_i \in F_i,\theta \in Sols(\langle V_{i}, C_{i} \cup \theta_{r_{ij}} \rangle) \}$
is the set of ground formula of $\langle F_i,C_i \rangle$ corresponding to $\theta_{r_{ij}}$.

We define the edge set of $R_g$ as follows. For each lifted edge $\langle r_i, r_k \rangle \in R_l$
compute the set $E = \{ (v_{ij}, v_{kl} ) \mid$ $\langle X_{ij}, F_{ij} \rangle$
 $\in Labels(r_i),$ $\langle X_{kl}, F_{kl} \rangle$ $\in Labels(r_k)$,
$ X_{ij} \cap X_{kl} \neq \emptyset \}.$
The ground region graph $R_g$ is defined as the $3$-tuple $\langle V, E, L \rangle$, where
$L = \{ l_{ij} \mid \forall i, l_{ij} \in Labels(r_i) \}$ and $V = \{ v_{ij} \lvert \forall i, l_{ij} \in Labels(r_i)\}$.
A lifted region graph is valid if and only if its corresponding ground region graph is valid.

\begin{theorem}
Let $M$ be an MLN. Let $M_g=\langle X, F \rangle$ be the Markov network corresponding to $M$.
A lifted region graph $R_{l}$ is \emph{valid} w.r.t $M$ iff its corresponding ground region graph
$R_{g}=\langle V,E,L\rangle$ is \emph{valid} w.r.t. $M_{g}$. A ground region graph
is \emph{valid} if it obeys the \emph{running intersection} property, which states that $\forall v_1,v_2 \in V,
x \in l(v_1) \wedge x \in l(v_2) \rightarrow \exists v_3 \in V \ni x \in l(v_3) \wedge v_3 \in \mathcal{E}(v_1)
\wedge v_3 \in \mathcal{E}(v_2)$.
\end{theorem}

\subsection{Statistics over the Simulated Region Graph}
\label{statistics}

\begin{figure*}[t]
\center
\includegraphics[page=1,width=0.31\linewidth,trim=20 5 5 50,clip]{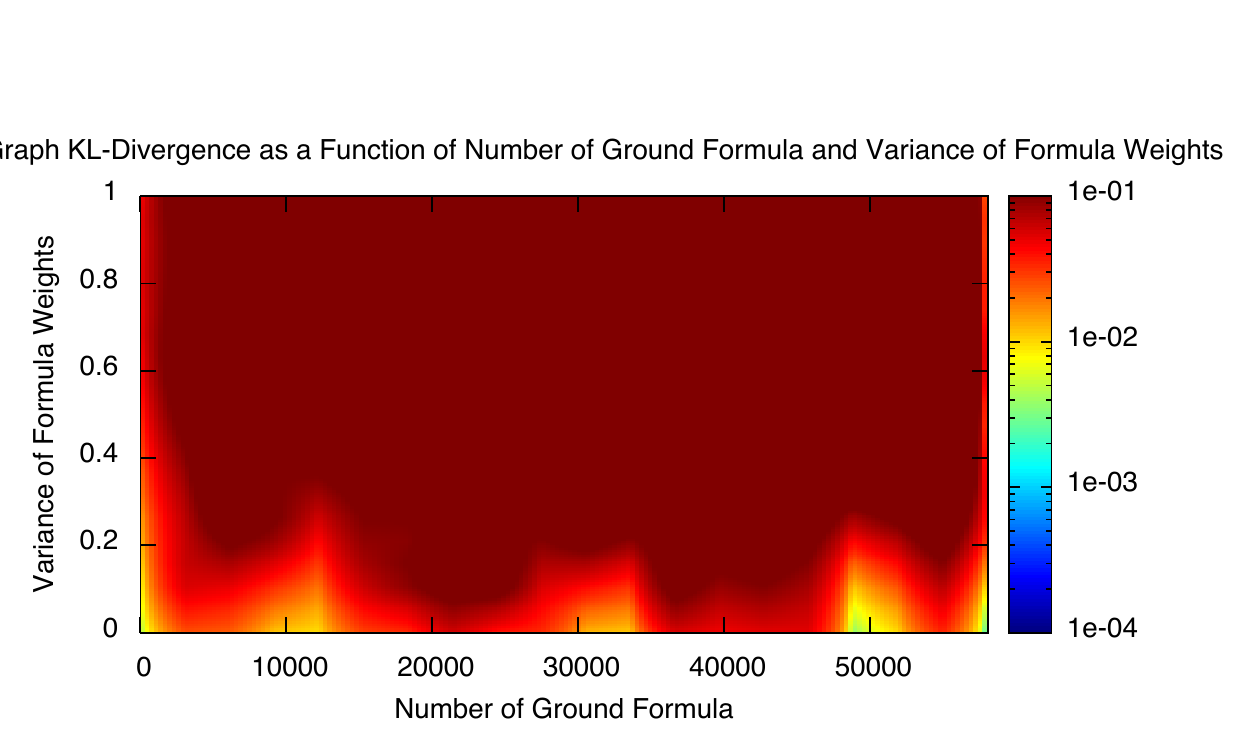}
\includegraphics[page=3,width=0.31\linewidth,trim=20 5 5 50,clip]{random-kldivergence.pdf}
\includegraphics[page=2,width=0.31\linewidth,trim=20 5 5 50,clip]{random-kldivergence.pdf}

\includegraphics[page=1,width=0.31\linewidth,trim=20 5 5 50,clip]{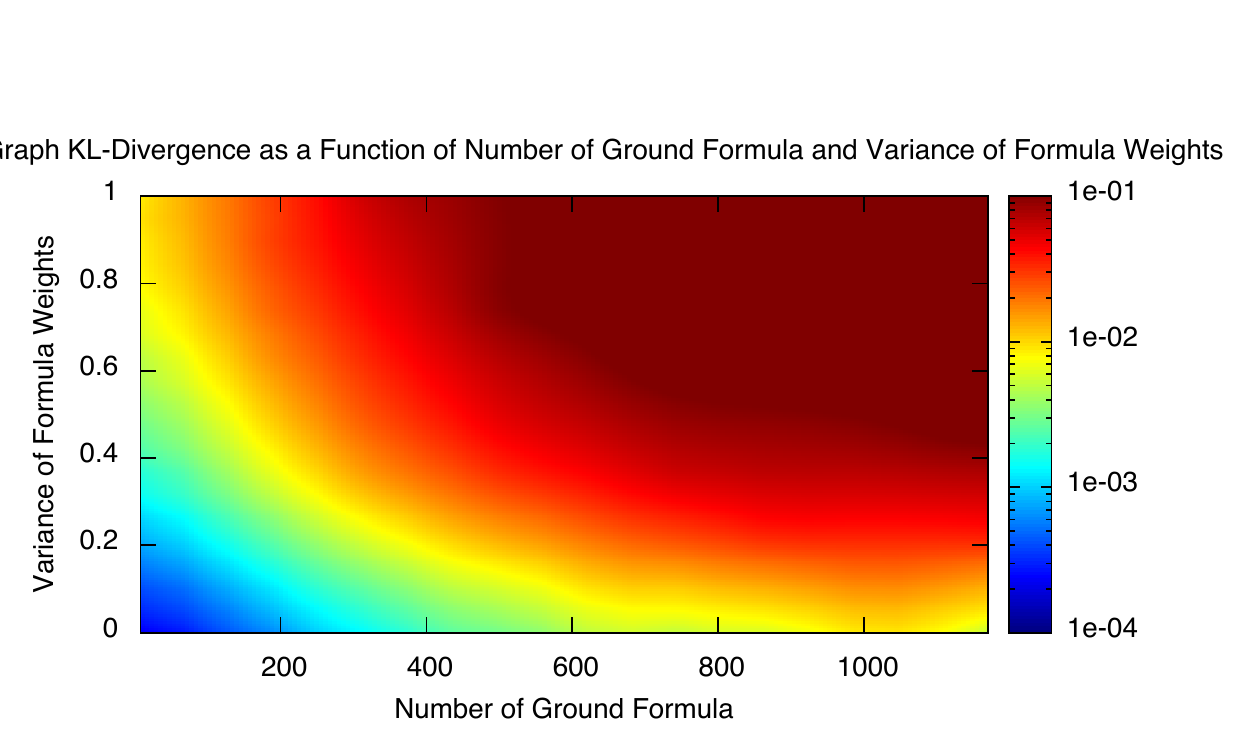}
\includegraphics[page=3,width=0.31\linewidth,trim=20 5 5 50,clip]{fs-kldivergence.pdf}
\includegraphics[page=2,width=0.31\linewidth,trim=20 5 5 50,clip]{fs-kldivergence.pdf}
\caption{Row 1: Random Tractable MLNs, Row 2: FSPC MLN, Left column: ground formula, ground messages,
Middle Column: lifted formula, ground messages, Right Column: lifted formula, lifted messages}
\label{resultsFigures}
\end{figure*}

The LGBP propagation algorithm
only requires statistics about the number of identical messages send during message passing. 
Specifically, the message-update rule requires the following quantities:
\begin{enumerate}[topsep=0pt,itemsep=-1ex,partopsep=1ex,parsep=1ex]
\item $G_P(r,r_p,R_l)$ - the number of copies of
$\langle r_p,r \rangle \in R_l$ directed into a single copy of $r$ from all copies of $r_p$ in $R_g$.
\item $G_D(r,r_d,R_l)$ - the number of copies of $r_d$ that are descendants of a single copy of $r$ in $R_g$.
\item $G_\mathcal{E}(r,r_d,r_{d_p})$ - Given lifted region nodes $r,r_d,r_{d_p}$ where:
(a) $r_d$ is a descendant of $r$ in $R_l$, (b) $r_{d_p}$ is a parent of $r_d$ in $R_l$,
(c) $v_r \in R_g$ is a single copy of $r$, and (d) $v_{d_r} \in R_g$ is a single copy of $r_d$,
$G_\mathcal{E}(r,r_d,r_{d_p})$ is the number of copies of $r_{d_p}$ in $R_g$
(excluding $v_r$) that are parents of $v_{d_r} \in R_g$ but not descendants of $v_r \in R_g$.
\end{enumerate}

Each of these quantities can be computed (via formulation as a CSP) without explicitly constructing the
ground region graph. We omit the derivation due to space constraints.

\subsection{Message Passing}
We present a lifted version of the parent-to-child algorithm. Each lifted region
$r$ has a belief given by $b_r(x_r) =$
\setlength{\belowdisplayskip}{4pt} \setlength{\belowdisplayshortskip}{4pt}
\setlength{\abovedisplayskip}{4pt} \setlength{\abovedisplayshortskip}{4pt}
\begin{multline}
\prod_{f_i \in M_r} f_i \left( \prod_{r_p \in P(r)} m_{r_p \rightarrow r}^{G_P (r, r_p, R_l)} \right)
\\
\left( \prod_{r_d \in D(r)}  \left( \prod_{ r'_p \in P(r_d)} m_{r'_p \rightarrow r_d}^{G_\mathcal{E}(r,r_d,r_{d_p},R_l)} \right)^{G_D(r,r_d,R_l)} \right)
\end{multline}

Here $P(r)$ is the set of lifted regions that are parents to lifted region $r$ and $D(r)$ is the set of all
lifted regions that are descendants of lifted region $r$.
The message-update rule for $m_{r_p \rightarrow r}$ is obtained by setting the beliefs at regions $r$ and $r_p$
to be equal over their message variables, and is given by $m_{r_p \rightarrow r}(x_R) =$
\begin{multline}
\frac{
\sum_{x_{P \setminus R}} b_p(x_r)
}
{
\left( \prod_{f_i \in M_r} f_i \right)
\left( m_{r_p \rightarrow r}^{G_P(r,r_p)-1} \right) \left( \prod_{r_p' \in P'(r,r_p)} m_{r_p' \rightarrow r}^{G_P(r,r_p') } \right)
}
\end{multline}
where $P'(r,r_p)$ is the set of lifted regions that are parents of $r$ in $R_l$ excluding $r_p$.

\subsection{Intra-Region Inference and Region Graph Construction}
Each message $m_{r_p \rightarrow r}(x_R)$ is computed in the parent region, $r_p =
\langle M_p, V_{pg}, E_{r_pg} \rangle$ by running inference over the lifted factorization
of a single grounding of $r_p$ given by $E_{r_pg}$. Inference is handled via any
exact lifted inference algorithm \cite{braz07,gogate&domingos11b,broeck&al11,smith&gogate2015}.

LGBP is a general method that works on any valid lifted region graph. A natural construction
method is to heuristically grouping formulas based on the cost of lifted inference and
then apply either (1) the variational cluster method \cite{kikuchi51} or
(2) a mini-bucket based scheme \cite{dechteretal2002}  over the intersections of efficiently
available sets of marginals.

\section{Related Work}
Lifting LBP relies on the observation that the factor graph structure
gives rise to message-level symmetries when applied to SRMs \cite{jaimovich&meshi2012}.
Both FOBP \cite{singla&domingos08} and Counting Belief Propagation \cite{kersting&al09}
propose algorithms to exploit these message-level
symmetries. FOBP presents an iterative algorithm for shattering a MLN and a set of
evidence into a lifted factor graph upon which messages are
split into groups guaranteed to be identical on every iteration. CBP compresses
a propositional factor graph by identifying identical
messages and lifting over them. LGBP differs from both of these
algorithms in that they perform the intra-cluster exact inference step on the
propositional level, while LBGP can exploit symmetries present in each region as well
as the structure of the messages being passed.

The Lifted RCR algorithm (LRCR) \cite{broecketal12} lifts the propositional
RCR algorithm \cite{choi&darwiche10}. The RCR
algorithm is a generalization of GBP
in which equality constraints between random variables in different potentials are relaxed,
these relaxations are compensated for (e.g. via message passing),
and then some constraints are recovered,
based on a heuristic. LRCR extends this framework to
lifted models. Like LGBP, LRCR uses lifted inference to allow dramatically
larger scopes at each region. However, LRCR still performs the `compensate` step by
passing messages over the marginals of single ground variables. LGBP goes one step
further; when possible it passes compact messages over the joint distribution of exchangeable
variables, thus yielding a region with fewer edges.

More recently, researchers have introduced symmetry-exploiting techniques
that permit formulation of  the approximate inference task as an efficient optimization problem 
These methods admit a reparameterization SRM inference over a reduced
variable space; the problem can then be solved by standard LP techniques for MAP inference
\cite{mladenovetal14} and by variational methods for marginal inference
\cite{bui&al14,mladenov&kersting15}.

\section{Experimental Results}

We conduct two sets of experiments. We focus on models which are amenable to exact inference so that
we can compare accuracy of different message passing structures.

\paragraph{Random Tractable Models}
We generated 1000 sets of 15 first-order clauses, $\{ KB_1, \ldots, KB_{1000} \}$. Each clause is
of the form $x \vee y \vee z$, where $x,y,z$ are randomly selected from the set $\{ R_1(x_1), \ldots, R_{15}(x_{15}) \}$.
For each $KB_i, $ variance $\sigma \in \{0.0, 0.1, \ldots, 1.0 \},$ and domain size $d \in \{1, \ldots, 20 \}$, we
generate an MLN by assigning the domain of all variables in $KB_i$ to $\{1, \ldots, d \}$ and assigning
each clause in $KB_i$ a weight sampled from $\mathcal{N}(0,\sigma)$.

For each randomly generated MLN, we construct three lifted region graphs. All region graphs place a single
lifted formula in each top level region. The first region graph grounds the top level formula and passes messages
over ground variables, similar to FOBP. The second
region graph builds a lifted factorization of all ground formulas in each top level cluster, but passes
messages over ground variables. The third region graph builds a lifted factorization of each cluster, and
communicates via joint messages over exchangeable atoms when the structure allows. For each model, we compute
the true marginals over each lifted atom, and then compute the KL-divergence of these (single variable)
marginals from those returned by LGBP. Figure \ref{resultsFigures}(top) shows
KL-divergence as a function of variance and domain size for each structure. 
The results show that the lifted region graph structure returns accurate results for a
significantly larger range of domain size and variance than either of the other structures.

\paragraph{Friends, Smokers, Parents, Cancer MLN Results}
The second experimental setup mirrors the first; however all 1000 runs of the algorithm are performed
on the same model, a complication of the Friends and Smokers MLN:
\setlength{\abovedisplayskip}{2pt}
\setlength{\belowdisplayskip}{2pt}
\scriptsize
\begin{gather*}
\langle Smokes(x) \wedge Friends(x,y) \rightarrow Smokes(y) \rangle \\
\langle Smokes(x) \rightarrow Cancer(x) \rangle \\
\langle Cancer(y) \wedge ParentOf(y,x) \rightarrow Cancer(x) \rangle \\
\langle Smokes(y) \wedge ParentOf(x,y) \rightarrow Smokes(x) \rangle
\end{gather*}
\normalsize
 
We also added formulas for each singleton atom. Again we randomly generated weights as per
the procedure detailed for random models, and we ran the algorithm 1000 times on the same three
types of region graphs. Figure \ref{resultsFigures}(bottom) shows KL-divergence as a function of variance
and domain size for each algorithm.
Figure \ref{resultsFigures} demonstrates that while FOBP can yield quite accurate results in some cases,
it is not resilient to large variance in formula weights, and that increasing the domain size can further
exacerbate its accuracy. We observed that FOBP region graph structure
generally takes more iterations than either of the other region graph structures, and often fails to converge
for even moderately diverse weights. Clustering groundings of the same formula offers a significant
improvement in both convergence and accuracy of the returned results. We observed that the addition of
joint message passing requires slightly more iterations, but will converge to superior results on a
wider variety of models.

\section{Conclusions and Future Work}
For message-passing based inference methods in PGMs, one strategy for realizing accurate
approximations is to reduce the number of edges in the message-passing structure. By exploiting
techniques for exact lifted inference, we have extended this strategy to SRMs. We have
presented a Lifted Generalized Belief Propagation algorithm and demonstrated that the algorithm
improves the overall accuracy of the approximation on a number of models.
For future work, our first goal is to develop a lifted region graph construction algorithm that
clusters formulas into top-level regions such that (1) the complexity of inference at each cluster is
bounded, and (2) the number of messages in the model is minimized. Second, we aim to employ the
LGBP algorithm for efficient weight learning over large and complicated models. Third,
we aim to generalize inference over the lifted region graph structure to algorithms using lifted variational
inference principles \cite{bui&al13}.

\vspace{-0.5em}
\subsubsection{Acknowledgements}
This research was funded by the Defense Advanced Research Projects Agency (DARPA) Probabilistic Programming for Advanced Machine Learning (PPAML) Program under Air Force Research Laboratory (AFRL) prime contract no. FA8750-14-C-0005.

\bibliography{all}
\bibliographystyle{aaai}
\end{document}

%% file: figExTree1.tikz
\begin{tikzpicture}[xscale=1.0, yscale=0.9]
	\begin{pgfonlayer}{nodelayer}
		\node [style=ornode] (0) at (-9.5, 2.75) {};
		\node [style=label] (1) at (-8.75, 2.75) {$R(C)$};
		\node [style=label] (2) at (-9, 2.25) {$[y]$};
		\node [style=ornode] (3) at (-9.5, 1.5) {};
		\node [style=label] (4) at (-8.75, 1.5) {$S(D)$};
		\node [style=label] (5) at (-5, 2.25) {$[]$};
		\node [style=label] (6) at (-4.75, 1.5) {$S(C)$};
		\node [style=ornode] (7) at (-5.25, 2.75) {};
		\node [style=ornode] (8) at (-5.25, 1.5) {};
		\node [style=label] (9) at (-4.75, 2.75) {$R(C)$};
		\node [style=label] (10) at (-5.75, 1.75) {$S$};
		\node [style=gandnode] (11) at (-7.5, 1.25) {};
		\node [style=gandnode] (12) at (-7, 1.25) {};
		\node [style=label] (13) at (-6, 2) {2};
		\node [style=label] (14) at (-7.75, 1.75) {$S$};
		\node [style=label] (15) at (-6.75, 2.75) {$R$};
		\node [style=ornode] (16) at (-6.25, 1.75) {};
		\node [style=gandnode] (17) at (-8.25, 2.25) {};
		\node [style=gandnode] (18) at (-7.25, 2.25) {};
		\node [style=gandnode] (19) at (-8, 1.25) {};
		\node [style=label] (20) at (-8, 2) {2};
		\node [style=label] (21) at (-6.75, 1.75) {$S$};
		\node [style=ornode] (22) at (-7.25, 2.75) {};
		\node [style=gandnode] (23) at (-7.25, 3.25) {};
		\node [style=gandnode] (24) at (-8.5, 1.25) {};
		\node [style=gandnode] (25) at (-6, 1.25) {};
		\node [style=ornode] (26) at (-7.25, 1.75) {};
		\node [style=gandnode] (27) at (-6.25, 2.25) {};
		\node [style=ornode] (28) at (-8.25, 1.75) {};
		\node [style=label] (29) at (-7, 2) {2};
		\node [style=gandnode] (30) at (-6.5, 1.25) {};
		\node [style=label] (31) at (-1.5, 1.75) {$S$};
		\node [style=gandnode] (32) at (-3, 2.25) {};
		\node [style=ornode] (33) at (-2, 1.75) {};
		\node [style=ornode] (34) at (-3, 2.75) {};
		\node [style=gandnode] (35) at (-2, 2.25) {};
		\node [style=gandnode] (36) at (-2.25, 1.25) {};
		\node [style=label] (37) at (-2.5, 1.75) {$S$};
		\node [style=label] (38) at (-2.5, 2.75) {$R$};
		\node [style=gandnode] (39) at (-4, 2.25) {};
		\node [style=ornode] (40) at (-4, 1.75) {};
		\node [style=label] (41) at (-3.5, 1.75) {$S$};
		\node [style=gandnode] (42) at (-2, 1.25) {};
		\node [style=gandnode] (43) at (-3, 3.25) {};
		\node [style=ornode] (44) at (-3, 1.75) {};
		\node [style=gandnode] (45) at (-1.75, 1.25) {};
		\node [style=gandnode] (46) at (-3.25, 1.25) {};
		\node [style=gandnode] (47) at (-3, 1.25) {};
		\node [style=gandnode] (48) at (-2.75, 1.25) {};
		\node [style=gandnode] (49) at (-4.25, 1.25) {};
		\node [style=gandnode] (50) at (-4, 1.25) {};
		\node [style=gandnode] (51) at (-3.75, 1.25) {};
	\end{pgfonlayer}
	\begin{pgfonlayer}{edgelayer}
		\draw [style=regionedge] (0) to (3);
		\draw [style=regionedge] (7) to (8);
		\draw [style=regionedge] (23) to (22);
		\draw [style=regionedge] (18) to (26);
		\draw [style=regionedge] (26) to (11);
		\draw [style=regionedge, in=90, out=270] (22) to (18);
		\draw [style=regionedge, in=124, out=-56] (26) to (12);
		\draw [style=regionedge] (27) to (16);
		\draw [style=regionedge] (16) to (30);
		\draw [style=regionedge] (16) to (25);
		\draw [style=regionedge] (17) to (28);
		\draw [style=regionedge] (28) to (24);
		\draw [style=regionedge] (28) to (19);
		\draw [style=regionedge] (22) to (17);
		\draw [style=regionedge] (22) to (27);
		\draw [style=regionedge] (43) to (34);
		\draw [style=regionedge] (32) to (44);
		\draw [style=regionedge, in=90, out=270] (34) to (32);
		\draw [style=regionedge] (35) to (33);
		\draw [style=regionedge] (33) to (36);
		\draw [style=regionedge] (33) to (42);
		\draw [style=regionedge] (39) to (40);
		\draw [style=regionedge] (34) to (39);
		\draw [style=regionedge] (34) to (35);
		\draw [style=regionedge] (33) to (45);
		\draw [style=regionedge] (40) to (49);
		\draw [style=regionedge] (50) to (40);
		\draw [style=regionedge] (40) to (51);
		\draw [style=regionedge] (44) to (46);
		\draw [style=regionedge] (47) to (44);
		\draw [style=regionedge] (44) to (48);
	\end{pgfonlayer}
\end{tikzpicture}

%% file: fig1.tikz
\begin{tikzpicture}
	\begin{pgfonlayer}{nodelayer}
		\node [style=variablenode] (0) at (-6.5, 1.25) {$S(1)$};
		\node [style=clusternode] (1) at (-5, 2.75) {$R(10)T(10)$};
		\node [style=clusternode] (2) at (-6.5, 2.75) {$R(1)S(1)$};
		\node [style=label] (3) at (-5.75, 1.25) {. . .};
		\node [style=label] (4) at (-5.25, 2.25) {. . .};
		\node [style=label] (5) at (-5.75, 2.75) {. . .};
		\node [style=label] (6) at (-6.25, 2.25) {. . .};
		\node [style=variablenode] (7) at (-5, 1.25) {$T(10)$};
		\node [style=label] (8) at (-3, 1.25) {. . .};
		\node [style=label] (9) at (-3.25, 1.75) {. . .};
		\node [style=variablenode] (10) at (-3.5, 1.25) {$R(1)$};
		\node [style=variablenode] (11) at (-2.25, 1.25) {$R(10)$};
		\node [style=label] (12) at (-2.5, 1.75) {. . .};
		\node [style=liftedclusternode] (13) at (-3.5, 2.75) {$R(x)S(y)$};
		\node [style=liftedclusternode] (14) at (-2.25, 2.75) {$R(x)T(z)$};
		\node [style=liftedvariablenode] (15) at (-0.25, 1.25) {$R(x)$};
		\node [style=liftedclusternode] (16) at (0.5, 2.75) {$R(x)T(z)$};
		\node [style=liftedclusternode] (17) at (-1, 2.75) {$R(x)S(y)$};
	\end{pgfonlayer}
	\begin{pgfonlayer}{edgelayer}
		\draw [style=regionedge] (2) to (0);
		\draw [style=regionedge] (1) to (7);
		\draw [style=regionedge] (0) to (5);
		\draw [style=regionedge] (0) to (6);
		\draw [style=regionedge] (7) to (4);
		\draw [style=pathedge] (13) to (10);
		\draw [style=pathedge] (13) to (11);
		\draw [style=pathedge] (14) to (11);
		\draw [style=pathedge] (14) to (10);
		\draw [style=regionedge] (13) to (9);
		\draw [style=regionedge] (14) to (12);
		\draw [style=pathedge] (17) to (15);
		\draw [style=pathedge] (16) to (15);
		\draw [style=regionedge] (7) to (5);
	\end{pgfonlayer}
\end{tikzpicture}

%% file: figRST.tikz
\begin{tikzpicture}
	\begin{pgfonlayer}{nodelayer}
		\node [style=liftedclusternode] (0) at (-11.5, 6.25) {$R(x),S(y)$};
		\node [style=liftedclusternode] (1) at (-5.75, 6.25) {$R(x),T(z)$};
		\node [style=liftedclusternode] (2) at (-8.5, 6.25) {$S(y),T(z)$};
		\node [style=label] (3) at (-10, 6) {$S$(D)};
		\node [style=ornode] (4) at (-10.5, 6.5) {};
		\node [style=label] (5) at (-10.25, 6.25) {$[y]$};
		\node [style=ornode] (6) at (-10.5, 6) {};
		\node [style=label] (7) at (-10, 6.5) {$R$(C)};
		\node [style=label] (8) at (-4.5, 6.25) {$[z]$};
		\node [style=ornode] (9) at (-4.75, 6.5) {};
		\node [style=label] (10) at (-4.25, 6) {$T$(D)};
		\node [style=ornode] (11) at (-4.75, 6) {};
		\node [style=label] (12) at (-4.25, 6.5) {$R$(C)};
		\node [style=label] (13) at (-7, 6.5) {$S$(C)};
		\node [style=label] (14) at (-7, 6) {$T$(D)};
		\node [style=ornode] (15) at (-7.5, 6.5) {};
		\node [style=ornode] (16) at (-7.5, 6) {};
		\node [style=label] (17) at (-7.25, 6.25) {$[z]$};
		\node [style=ornode] (18) at (-9, 5.5) {};
		\node [style=label] (19) at (-8.5, 5.5) {$R$(C)};
		\node [style=liftedclusternode] (20) at (-9.75, 5.5) {$R(x)$};
		\node [style=ornode] (21) at (-4.75, 5.5) {};
		\node [style=liftedclusternode] (22) at (-5.5, 5.5) {$T(z)$};
		\node [style=label] (23) at (-4.25, 5.5) {$T$(G)};
		\node [style=liftedclusternode] (24) at (-11.5, 5.5) {$S(y)$};
		\node [style=ornode] (25) at (-11, 5.5) {};
		\node [style=label] (26) at (-10.5, 5.5) {$S$(G)};
		\node [style=ornode] (27) at (-7, 5.5) {};
		\node [style=label] (28) at (-6.5, 5.5) {$S$(C)};
		\node [style=liftedclusternode] (29) at (-7.75, 5.5) {$S(y)$};
		\node [style=ornode] (30) at (-9, 5) {};
		\node [style=liftedclusternode] (31) at (-9.75, 5) {$S(y)$};
		\node [style=label] (32) at (-8.5, 5) {$S$(G)};
	\end{pgfonlayer}
	\begin{pgfonlayer}{edgelayer}
		\draw [style=regionedge] (4) to (6);
		\draw [style=regionedge] (9) to (11);
		\draw [style=regionedge] (15) to (16);
		\draw [style=regionedge] (0) to (20);
		\draw [style=regionedge] (1) to (20);
		\draw [style=regionedge] (0) to (24);
		\draw [style=regionedge] (2) to (29);
		\draw [style=regionedge] (2) to (22);
		\draw [style=regionedge] (1) to (22);
		\draw [style=regionedge] (29) to (31);
		\draw [style=regionedge] (24) to (31);
	\end{pgfonlayer}
\end{tikzpicture}

%% file: figRSTground.tikz
\begin{tikzpicture}
	\begin{pgfonlayer}{nodelayer}
		\node [style=liftedclusternode] (0) at (-7.75, 2.5) {$S(1),S(2),T(1),T(2)$};
		\node [style=liftedclusternode] (1) at (-7, 1.75) {$S(1),S(2)$};
		\node [style=liftedclusternode] (2) at (-8.75, 1.75) {$R(1),R(2)$};
		\node [style=liftedclusternode] (3) at (-5, 2.5) {$R(1),R(2),T(1),T(2)$};
		\node [style=liftedclusternode] (4) at (-11.25, 1.75) {$S(1)$};
		\node [style=liftedclusternode] (5) at (-10.5, 2.5) {$R(1),R(2),S(1),S(2)$};
		\node [style=liftedclusternode] (6) at (-9, 1) {$S(1)$};
		\node [style=liftedclusternode] (7) at (-5, 1.75) {$T(1)$};
		\node [style=liftedclusternode] (8) at (-10.25, 1.75) {$S(2)$};
		\node [style=liftedclusternode] (9) at (-7.75, 1) {$S(2)$};
		\node [style=liftedclusternode] (10) at (-4, 1.75) {$T(2)$};
	\end{pgfonlayer}
	\begin{pgfonlayer}{edgelayer}
		\draw [style=regionedge] (5) to (2);
		\draw [style=regionedge] (3) to (2);
		\draw [style=regionedge] (5) to (4);
		\draw [style=regionedge] (0) to (1);
		\draw [style=regionedge] (0) to (7);
		\draw [style=regionedge] (3) to (7);
		\draw [style=regionedge] (1) to (6);
		\draw [style=regionedge] (4) to (6);
		\draw [style=regionedge] (8) to (9);
		\draw [style=regionedge] (9) to (1);
		\draw [style=regionedge] (0) to (10);
		\draw [style=regionedge] (10) to (3);
		\draw [style=regionedge] (8) to (5);
	\end{pgfonlayer}
\end{tikzpicture}